\documentclass{article}

\usepackage[preprint]{icml2026}

\usepackage{kotex}
\usepackage{microtype}
\usepackage{graphicx}
\usepackage{subfigure}
\usepackage{booktabs} 

\usepackage[utf8]{inputenc} 
\usepackage[T1]{fontenc}    
\usepackage{hyperref}       
\usepackage{url}            
\usepackage{booktabs}       
\usepackage{amsfonts}       
\usepackage{nicefrac}       
\usepackage{microtype}      
\usepackage{graphicx}
\usepackage{xcolor}         
\usepackage{amsmath}        
\usepackage{amssymb}
\usepackage{mathtools}
\usepackage{amsthm}
\usepackage{enumitem}
\usepackage{tikz}
\usepackage{xparse}

\usepackage{algorithm}
\usepackage{algorithmic}

\usepackage{soul}
\usepackage{expl3}

\ExplSyntaxOn
\NewDocumentCommand{\CapitalizeFirst}{m}
 {
  \text_titlecase_all:n { #1 }
 }
\ExplSyntaxOff

\newcommand*\circled[1]{\tikz[baseline=(char.base)]{
            \node[shape=circle,draw,fill=black,text=white,font=\sffamily\bfseries\footnotesize, inner sep=1.0pt] (char) {#1};}}

\newcommand{\reasoningreq}{reasoning request}
\newcommand{\genreq}{LLM completion request}
\newcommand{\pexit}{positive early exit}
\newcommand{\Pexit}{Positive early exit}
\newcommand{\nexit}{negative early exit}

\theoremstyle{plain}

\theoremstyle{definition}

\theoremstyle{remark}

\icmltitlerunning{Adaptive Parallel Monte Carlo Tree Search for Efficient Test-time Compute Scaling}

\begin{document}

\twocolumn[
  \icmltitle{Adaptive Parallel Monte Carlo Tree Search for Efficient Test-time Compute Scaling}

\icmlsetsymbol{equal}{*}

  \begin{icmlauthorlist}
    \icmlauthor{Hongbeen Kim}{kaist}
    \icmlauthor{Juhyun Lee}{kaist}
    \icmlauthor{Sanghyeon Lee}{kaist}
    \icmlauthor{Kwanghoon Choi}{kaist}
    \icmlauthor{Jaehyuk Huh}{kaist}

  \end{icmlauthorlist}

  \icmlaffiliation{kaist}{School of Computing, Korea Advanced Institute of Science and Technology, Daejeon, South Korea}

  \icmlcorrespondingauthor{Jaehyuk Huh}{jhhuh@casys.kaist.ac.kr}

  \icmlkeywords{Machine Learning, ICML}

  \vskip 0.3in
]

\printAffiliationsAndNotice{}

\begin{abstract}

Monte Carlo Tree Search (MCTS) is an effective test-time compute scaling (TTCS) method for improving the reasoning performance of large language models, but its highly variable execution time leads to severe long-tail latency in practice. Existing optimizations such as positive early exit, reduce latency in favorable cases but are less effective when search continues without meaningful progress.
We introduce {\it negative early exit}, which prunes unproductive MCTS trajectories, and an {\it adaptive boosting mechanism} that reallocates reclaimed computation to reduce resource contention among concurrent searches. Integrated into vLLM, these techniques substantially reduce p99 end-to-end latency while improving throughput and maintaining reasoning accuracy.

\end{abstract}

\section{Introduction}
Large language models (LLMs) have demonstrated strong performance across a wide range of tasks, but complex reasoning often benefits from allocating additional computation at inference time. Test-time compute scaling (TTCS) addresses this need by increasing inference-time computation through generation, search, and evaluation, enabling LLMs to solve more challenging reasoning tasks with higher accuracy~\cite{lightman2024verify-step, sun2024easytohard, zhao2025genprm, peng2025graphprm, setlur2025rewarding-progress}.

Within TTCS, the choice of search strategy plays a central role in determining reasoning performance. Among existing approaches, Monte Carlo Tree Search (MCTS) has emerged as one of the most effective methods for LLM-based reasoning~\cite{hao2023rap, feng2023tsllm, guan2025rstar_math, misaki2025ab_mcts, zhou2024lats, antoniades2024swe_search, shi2025mc_dml, ma2024repounderstander}. When paired with a verifier or process reward model, MCTS consistently achieves higher accuracy than parallel candidate expansion methods (e.g., Best-of-N approaches). This advantage stems from its ability to balance exploration and exploitation, maintain multiple reasoning trajectories, and revisit paths with low early evaluations when their long-term potential remains promising.

Despite its accuracy advantages, MCTS poses a significant challenge in practical serving environments due to its inherently sequential execution. As the search progresses, decisions increasingly depend on prior explorations, causing execution time to grow rapidly with search depth. As a result, a small number of compute-intensive requests can disproportionately consume system resources, leading to pronounced long-tail latency. Moreover, deeper search does not always translate into better reasoning outcomes: MCTS may continue consuming substantial computation even when further exploration provides little benefit, fundamentally constraining its deployability in production settings.

Prior work has primarily explored two strategies to mitigate MCTS latency~\cite{fu2024certaindex, liu2018watch}. First, although not applied to TTCS, parallel MCTS variants relax strict serial dependencies by allowing multiple rollouts to proceed asynchronously, enabling concurrent expansion and reducing the wall-clock time of individual searches~\cite{liu2018watch}. Second, positive early exit techniques terminate the search once intermediate evaluations surpass a predefined confidence threshold, avoiding unnecessary computation when a high-quality solution is found early~\cite{fu2024certaindex}. These approaches are particularly effective when the search converges quickly or when early evaluations provide reliable guidance.

However, in realistic inference workloads, tail latency cannot be eliminated by these techniques alone. In many cases, search continues without clear signs of improvement despite additional exploration. Without a mechanism to identify and stop such unproductive searches, MCTS persists until completion, expending substantial computation for little benefit. While parallelization may reduce the execution time of individual searches, it does not reduce total computation. In capacity-constrained serving environments, this sustained overhead increases queueing delay and further exacerbates tail latency. Positive early exit is similarly ineffective in these scenarios, as the search never reaches a sufficiently confident intermediate state.

In this work, we take a system-centric perspective on MCTS-based reasoning, reconceptualizing it as a system that requires active resource management rather than a search process to be merely accelerated. Building on this view, we introduce {\it negative early exit}, a mechanism that identifies and prunes search trajectories that continue to consume computation without improving solution quality. In addition, we propose an {\it adaptive boosting mechanism} that redistributes reclaimed computational resources to concurrent searches with higher potential, prioritizing system capacity toward the most promising reasoning trajectories. Unlike the prior parallel MCTS~\cite{liu2018watch}, it controls parallelism dynamically based on the system state.

We integrate these ideas into a system that preserves the accuracy benefits of MCTS while substantially reducing long-tail latency and improving throughput. Built on vLLM~\cite{kwon2023pagedattention} and evaluated on Qwen-2.5~\cite{qwen2_5} and Llama-3.1~\cite{llama3}, it achieves up to 2.83× lower p99 end-to-end latency than serial MCTS and up to 1.46× lower latency than systems using positive early exit alone, while increasing throughput by up to 2.44× without sacrificing reasoning accuracy.

\section{Background}
\label{sect:background}


\subsection{Test-time Compute Scaling} 
The standard auto-regressive inference in Large Language Models (LLMs) often falls short on complex tasks that require deep exploration or reasoning.
Here, reasoning entails exploring alternatives, verifying intermediate steps, and iteratively refining a candidate before committing to an answer. Test-Time Compute Scaling (TTCS) provides this capability to systematically evaluate multiple candidate solutions by allocating extra inference compute and adjusting the computational effort for each prompt, which improves robustness and accuracy.

Snell et al.~\cite{snell2025scaling} analyze the trade-off between pre-training and test-time compute, proposing a compute-optimal scaling strategy tailored to individual prompts. They observe that while simple iterative refinements may suffice for straightforward tasks, complex reasoning increasingly requires verifier-guided search methods such as Monte Carlo Tree Search (MCTS).
Such search methods rely on an external verifier to evaluate intermediate reasoning steps and guide exploration toward promising trajectories.
A widely used verifier is the Process Reward Model (PRM), which generally produce a reward that reflects the local step correctness (i.e., how logically valid the current step is)~\cite{lightman2024verify-step, sun2024easytohard, zhao2025genprm, peng2025graphprm}.


\subsection{Structural Differences in Search Methods} 
\begin{figure}[t]
    \centering
    \includegraphics[width=1.0\linewidth]{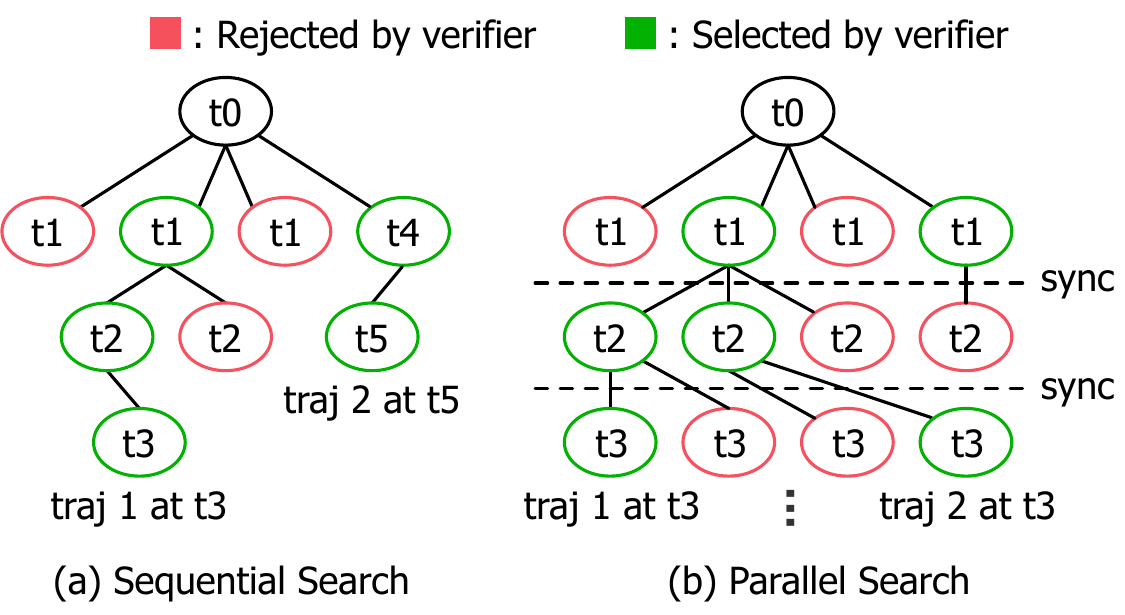}
    \caption{Two categories of the TTCS method: (a) sequential search and (b) parallel search. \textbf{t} denotes the time.}
    \label{fig:background_search_types}
\end{figure}

We categorize search methods into two distinct categories based on their trajectory generation patterns, as illustrated in Figure~\ref{fig:background_search_types}. 
The first category, which we term \textit{Sequential Search}, constructs trajectories sequentially through iterative execution.
MCTS~\cite{Kocsis2006BanditBasedMCP} is a representative example of Sequential Search, sequentially generating \textbf{trajectories} as candidate solutions through successive \textbf{rollout} executions.
In contrast, the second category, which we term \textit{Parallel Search}, generates trajectories in parallel.
Beam Search and REBASE~\cite{wu2024inferencescalinglawsempirical} are prominent examples of Parallel Search. Specifically, Beam Search maintains a fixed number of top-$k$ partial hypotheses at each step, expanding them in parallel to identify the most probable complete trajectories.
These methods involve a synchronization at each search step to prune or select candidates across the child nodes. 
For more details on the specific mechanics of each search method, please refer to Appendix~\ref{sec:appendix_mcts_detail} and Appendix~\ref{sec:appendix_bs_detail}.

These methodologies diverge in their approach to navigating the candidate solution space, resulting in disparate computational efficiencies. Specifically, sequential search improves inference efficiency by leveraging outcomes from prior executions to guide path selection, but its serial execution inherently restricts parallelism and slows down runtime. In contrast, parallel search methods generate a fixed number of samples at each step, incurring unnecessary computation, but enabling faster execution through parallelization. 
Consequently, sequential search supports more efficient but slower inference, whereas parallel search enables faster inference despite reduced computational efficiency.

\section{Motivation}
\label{sect:motivation}


\subsection{Search Efficiency of TTCS Methods} 
\begin{figure}[t]
    \centering
    \includegraphics[width=1.0\linewidth]{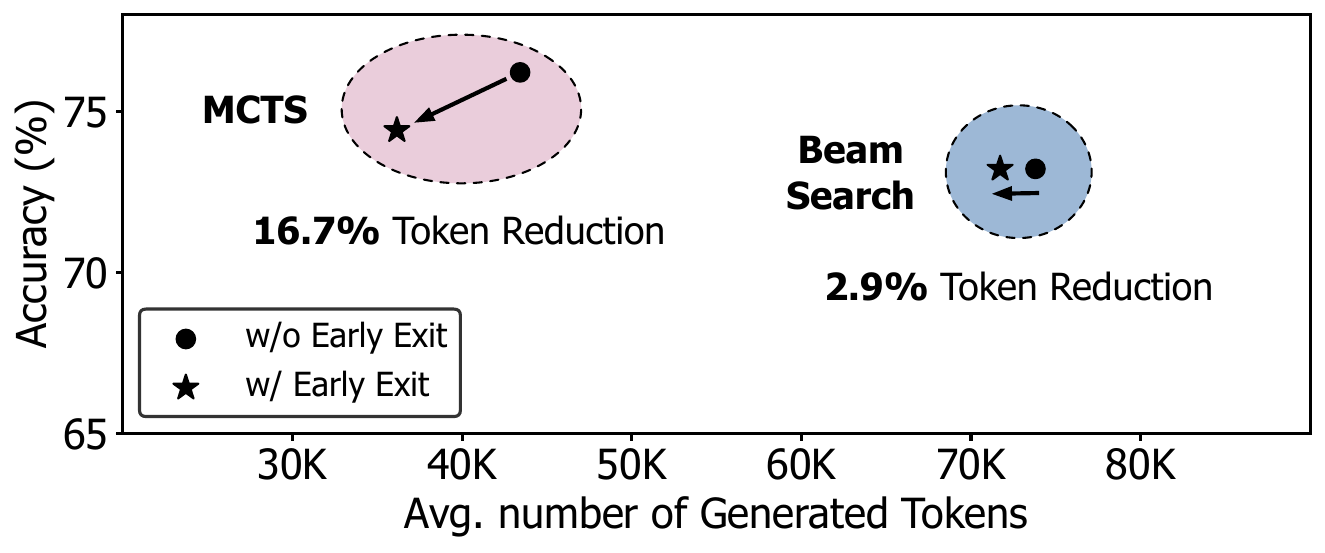}
    \caption{Comparison of search efficiency between Beam Search and MCTS, and the token reduction enabled by early exit, with MCTS using 12 rollouts and Beam Search using 8 beams.}
    \label{fig:motiv_search_efficiency}
\end{figure}

MCTS and Beam Search are widely used TTCS methods, they exhibit markedly different token generation efficiencies. MCTS constructs trajectories sequentially by repeatedly selecting the promising nodes, which helps minimize unnecessary token generation. In contrast, Beam Search expands a fixed number of candidates at each reasoning step, inevitably producing redundant or low-value tokens alongside useful ones. Consequently, even when achieving comparable accuracy, Beam Search tends to generate more tokens than MCTS, as also evidenced in Figure~\ref{fig:motiv_search_efficiency}

As shown in Figure~\ref{fig:motiv_search_efficiency}, MCTS generates fewer tokens on average compared to Beam Search while achieving slightly higher accuracy.
The efficiency gap between MCTS and Beam Search becomes significantly more pronounced when \textit{\pexit} is applied. Under this mechanism, the search process terminates as soon as it discovers a solution that satisfies a predefined quality threshold, thereby avoiding further unnecessary rollouts.
Figure~\ref{fig:motiv_search_efficiency} indicates that the token reduction achieved by \pexit is more pronounced for MCTS than for Beam Search, which can be attributed to the following factors.
In parallel search, the system must maintain a constant beam width $k$; thus, even if an optimal solution is identified at step $t$, the computational cost for the other $k-1$ candidates has already been incurred across all preceding steps. 
Conversely, sequential search methods explore the state space through iterative simulations, allowing them to bypass a vast majority of the potential search tree if a high-quality solution is found in an early iteration. 
This structural advantage allows sequential search to transform from an exhaustive exploration tool into a surgical retrieval mechanism, resulting in disproportionately higher token savings and superior scaling efficiency compared to the rigid, breadth-first nature of parallel methods.

\subsection{Latency Challenges of MCTS-Based TTCS} 

\label{subsect:tail-latency}

\begin{figure}[t]
    \centering
    \includegraphics[width=1.0\linewidth]{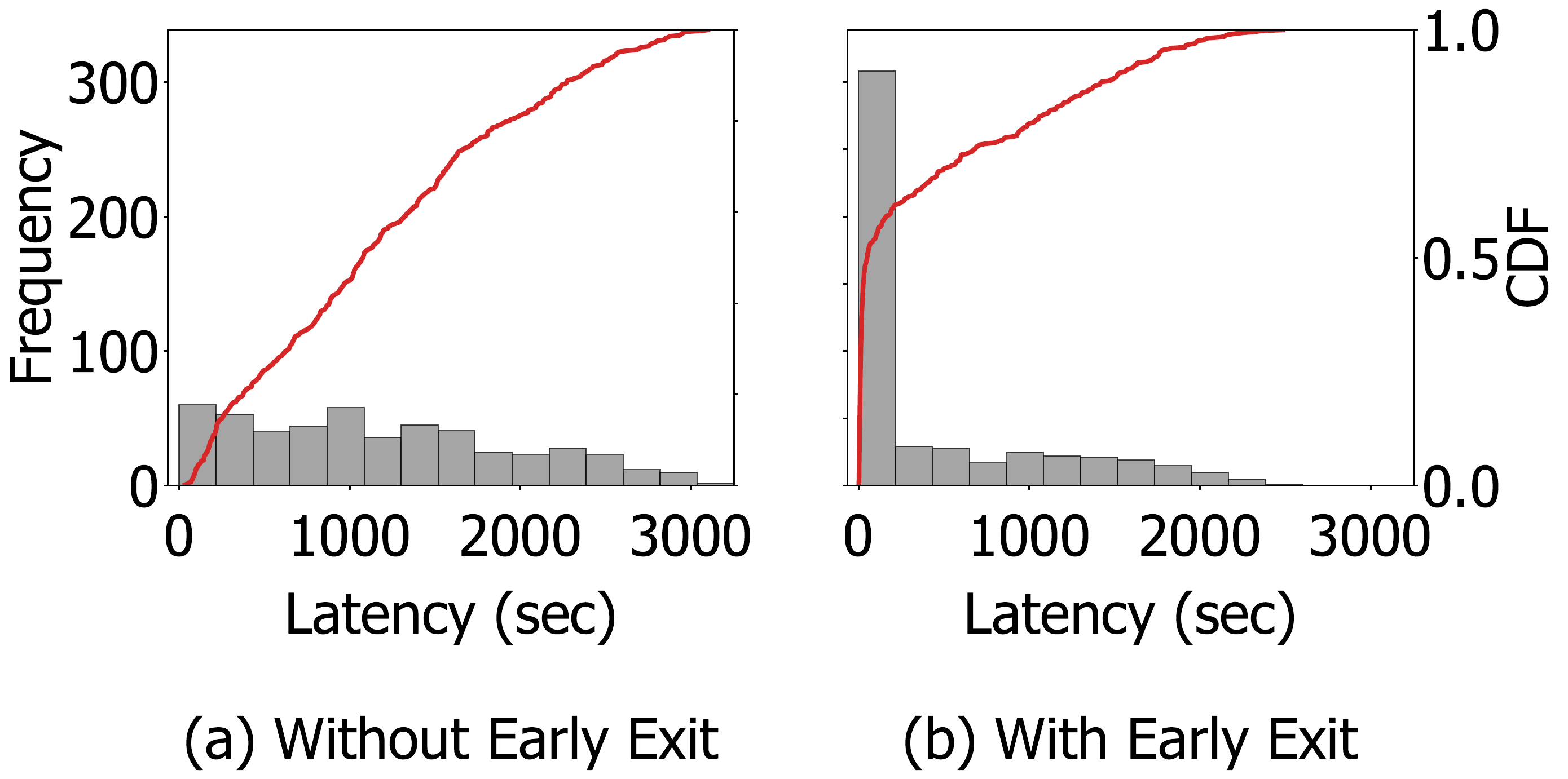}
    \caption{End-to-end latency distribution of sequential MCTS (a) with and (b) without early exit. Maximum number of rollouts is 32.}
    \label{fig:motiv_1-earlyexit}
\end{figure}

A major challenge in deploying MCTS-based reasoning is its long and unpredictable latency. 
The core reason is that this workflow imposes sequential dependencies—on step tokens within a rollout and on trajectory tokens across rollouts—forcing the model to generate many tokens strictly in sequence. 
Within a rollout the next step tokens depend on the preceding ones and are generated autoregressively. Each rollout must be executed using the statistics accumulated from previous rollouts, which causes execution across rollouts to proceed in a serial manner. 
This sequential dependency prevents parallelization across rollouts, making search execution intrinsically slower than parallel search. Figure~\ref{fig:motiv_1-earlyexit}(a) shows that sequential dependencies often lead to long latency.
As the rollout budget grows, this lack of parallelism between rollouts directly manifests as increased end-to-end latency and highly variable response times. {\Pexit} can be employed to address such end-to-end latency inflation. For example, Certaindex demonstrated that early-exit can substantially reduce the latency of many {\reasoningreq}s once the outcome is essentially determined~\cite{fu2024certaindex}. Overall, this adaptive allocation improves average-case performance and ensures that simpler queries complete quickly.

However, {\pexit} is not a complete solution. A non-trivial fraction of {\reasoningreq}s still fail to satisfy the exit condition, which forces the system to carry out the full set of rollouts As shown in Figure~\ref{fig:motiv_1-earlyexit}(b), these {\reasoningreq}s dominate the tail of the latency distribution and lead to rare but severe delays. For latency-sensitive applications, such long-tailed behavior is as detrimental as high mean latency because it undermines service predictability. Thus, while early-exit strategies mark progress, they leave open the question of how to systematically reduce tail latency in MCTS-based reasoning without sacrificing quality.


\subsection{Effect of Parallelization on MCTS-based TTCS}

\begin{figure}[t]
    \centering
\includegraphics[width=1.0\linewidth]{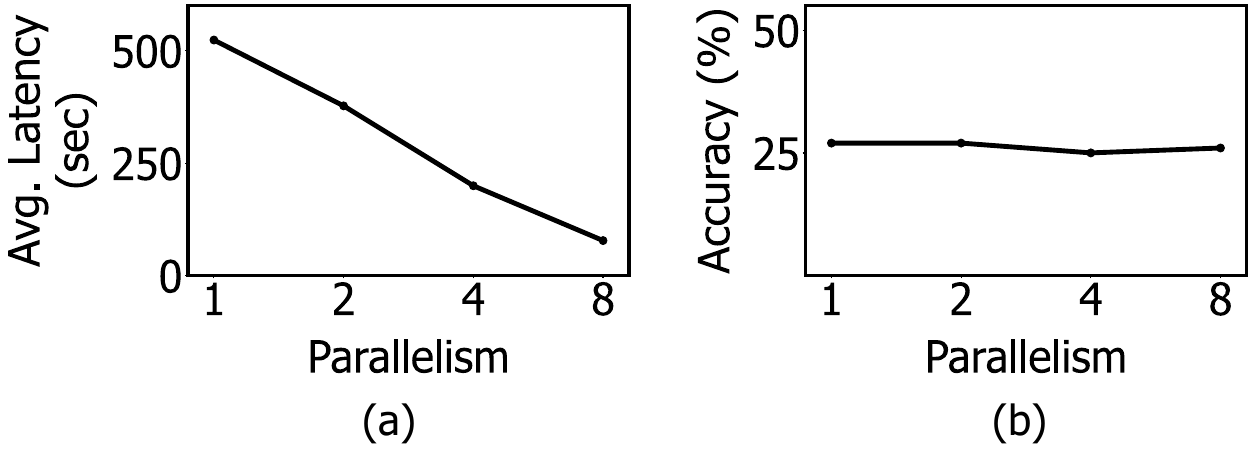}
\caption{Impact of parallelism on latency and accuracy. (a) Average latency per tree search decreases as parallelism increases. (b) Higher parallelism leads to minor fluctuations in accuracy. Results are reported on a subset of challenging problems that require generating a large number of tokens.}
\label{fig:motiv_effect_of_parallelism}
\end{figure}

While the parallelization mechanism  was not originally designed for TTCS, it is readily applicable to MCTS-based TTCS and yields significant latency reduction.
Figure~\ref{fig:motiv_effect_of_parallelism} illustrates the effectiveness of WU-UCT parallelization when applied to 100 problems sampled from the Math500 dataset, specifically selecting those that generated the highest total number of tokens during the vanilla MCTS. 
To prevent performance interference caused by resource contention, we parallelized each tree search sequentially, ensuring that only one tree search is processed at a time.
As shown in (a), the average latency per tree search decreases as the degree of parallelism increases. 
The reason it does not continue to decrease linearly is that parallelism is limited near the root node in the early stages of the search.
As illustrated in (b), accuracy is slightly affected as parallelism increases. 
We attribute this either to semantic degradation arising from the omission of certain statistics essential for exact MCTS, or to sampling variations caused by the floating-point precision limits of the language model. 
The overall low accuracy results from our selection of challenging instances that required a high volume of generated tokens.
The end-to-end accuracy evaluation is presented in Section~\ref{subsec:accuracy}.

\subsection{Opportunity of \CapitalizeFirst{\nexit}}
\label{subsec:motiv_negative_exit}

\begin{figure}[t]
    \centering
    \includegraphics[width=1.0\linewidth]{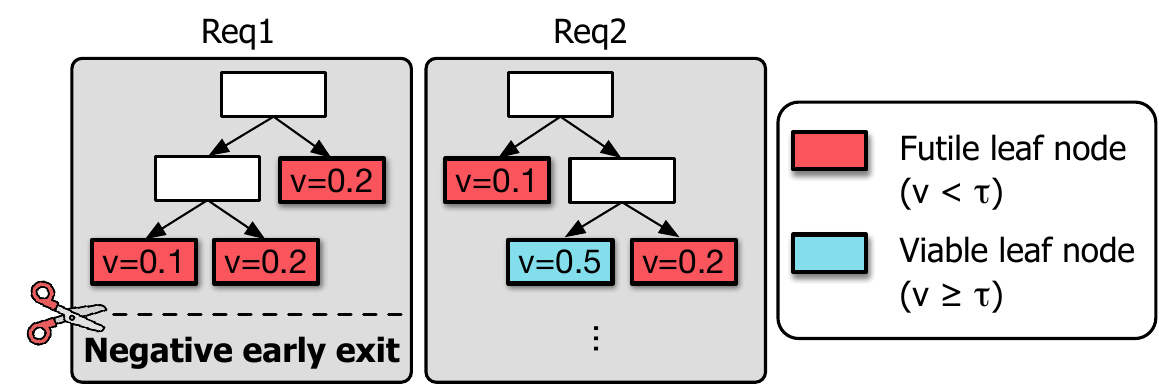}
    \caption{Overview of {\nexit} mechanism with a acceptance threshold of $\tau=0.3$.}
    \label{fig:motiv_ne_principle}
\end{figure}

Even with \pexit, tail latency can still remain high for requests that do not satisfy the early-exit condition. While parallelism can help mitigate this issue, it primarily reduces overall latency rather than specifically addressing tail latency. Therefore, it is crucial to identify the scenarios that lead to tail latency and to develop targeted solutions that eliminate unnecessary execution.

Tail latency is driven by two cases in which the problem is hard and {\pexit} cannot be triggered: (i) additional rollouts eventually discover a better solution; (ii) each rollout produces persistently low scores, leaving the problem unsolved within the rollout budget. For the case (ii), exhausting the rollout budget offers little value, since further rollouts are unlikely to produce the correct answer. 

To introduce {\nexit}, it is necessary to establish a criterion to assess whether additional rollouts are likely to discover a better trajectory. In MCTS, new trajectories are generated by selecting one of the current leaf nodes for expansion. The trajectory score is computed by aggregating the PRM reward scores of the nodes along the trajectory. Four common aggregation schemes are considered: (1) \textit{minimum}~\cite{wang2024mathshepherd, eurusprm-stage1}, (2) \textit{cumulative product}~\cite{zhang2025lqwen2.5-math}, (3) \textit{cumulative sum}~\cite{eurusprm-stage2}, or (4) \textit{average}~\cite{skywork, rlhflow}. Among them, the state-of-the-art PRM~\cite{qwen2_5_prm} argues that trajectory scoring based on (1) minimum or (2) cumulative product is most suitable. Accordingly, it adopts the cumulative product scheme.

Under both aggregation schemes, the resulting trajectory score is upper-bounded by the score of its leaf node; that is, the trajectory score can never exceed the leaf score. Therefore, leaf nodes below the acceptance threshold are termed \textit{futile}, while those above it are termed \textit{viable}. Based on this property, we propose the {\nexit} mechanism illustrated in Figure~\ref{fig:motiv_ne_principle}: when all leaf nodes in the current search tree are \textit{futile}, the request is early-exited. This corresponds to case (ii), where any future trajectory generated from the tree is guaranteed to have a score below the threshold, making further rollouts unlikely to yield a higher-quality solution. Conversely, if at least one leaf node is \textit{viable}, we continue the search, as additional rollouts may still discover a better trajectory.
\section{Method}
\label{sect:design}

\subsection{Overview}
Existing MCTS parallelization techniques like WU-UCT (§\ref{subsect:wu-uct}) offer a way to run multiple rollouts concurrently, but they do not adaptively manage the degree of parallelism. This is a critical limitation for LLM-based TTCS, where uncontrolled parallelism can saturate the inference engine and severely increase latency. We address this with our system, a scheduling framework that treats parallel rollouts not as a resource to maximize, but as a constrained computational budget—defined by the maximum number of concurrent rollouts—to be allocated efficiently and purposefully.

To ensure this budget is used effectively, our system first establishes an effective parallel search foundation with our WU-PUCT policy, which adapts the principle of unobserved counts to the PUCT selection rule to guide parallel workers toward diverse parts of the search tree. This is complemented by two novel intra-task pruning mechanisms: a probabilistic model that terminates rollouts with a low statistical likelihood of success, and a deterministic rule that stops them once their cumulative reward can no longer reach a target threshold. These mechanisms work together to minimize wasted work within each parallel search.

These intra-task optimizations are managed by a global, budget-aware scheduling policy that operates at the macro-level. A lightweight predictor helps the scheduler identify which tasks are likely to be long-running, allowing it to proactively increase their target parallelism level. Crucially, when an early-exit mechanism terminates a rollout, the scheduler can immediately reassign that freed-up budget slot to another task. This combination of efficient parallelization, fine-grained pruning, and dynamic, predictive resource allocation allows our system to reduce tail latency for complex tasks while keeping the total computation within the budget, thereby ensuring stable performance from the LLM serving infrastructure.

\subsection{Parallel MCTS with Partial Statistics}
\label{subsect:wu-uct}

Parallelizing MCTS remains a significant challenge due to the sequential nature of the search process, where each trajectory selection depends on up-to-date node statistics. Even before MCTS was widely adopted in reasoning tasks, its serial execution was a major computational bottleneck in classical settings. Without proper coordination, naive parallelization can degrade performance and lead to critical issues such as \textit{exploration collapse}—where multiple workers redundantly sample the same path—or \textit{exploitation failure}—where the search fails to converge on optimal nodes due to stale information~\cite{chaslot2008parallel, segal2010scalability, liu2018watch}.

To address these limitations, we extend our framework with a parallel rollout mechanism that integrates the principle of unobserved counts, originally introduced in the WU-UCT scheme~\cite{liu2018watch}. Unlike standard MCTS which enforces strict serial execution, this approach adopts a more relaxed strategy by allowing rollouts to continue using "in-flight" statistics when dependent updates have not yet been backpropagated. This enables a higher degree of parallelism and provides substantial speedup with minimal impact on search quality.

Specifically, we adapt the AlphaZero-style PUCT rule to this parallel setting by incorporating these unobserved counts into the selection criteria. We define modified visitation statistics as $\tilde{N}(s) = N(s) + O(s)$ and $\tilde{N}(s,a) = N(s,a) + O(s,a)$, where $O(\cdot)$ denotes the number of in-flight rollouts that have been launched but not yet completed. The resulting selection rule, which we term \textit{WU-PUCT}, is formulated as:
\[
\text{WU-PUCT}(s,a) = Q(s,a) + c_{\text{puct}} \cdot P(s,a) \cdot \frac{\sqrt{\tilde{N}(s)}}{1 + \tilde{N}(s,a)}.
\]
By incorporating ongoing rollouts into the node statistics, WU-PUCT effectively penalizes nodes that are already being explored by other workers. This preserves the exploration--exploitation balance and prevents redundant computation under high-throughput parallel execution. This mechanism is natively embedded into our \texttt{launch\_rollout()} API, allowing the scheduler to initiate multiple rollouts concurrently while maintaining consistency through the corrected selection rule.

\subsection{\CapitalizeFirst{\nexit{}}}
\begin{figure}[t]
    \centering
\includegraphics[width=1.0\linewidth]{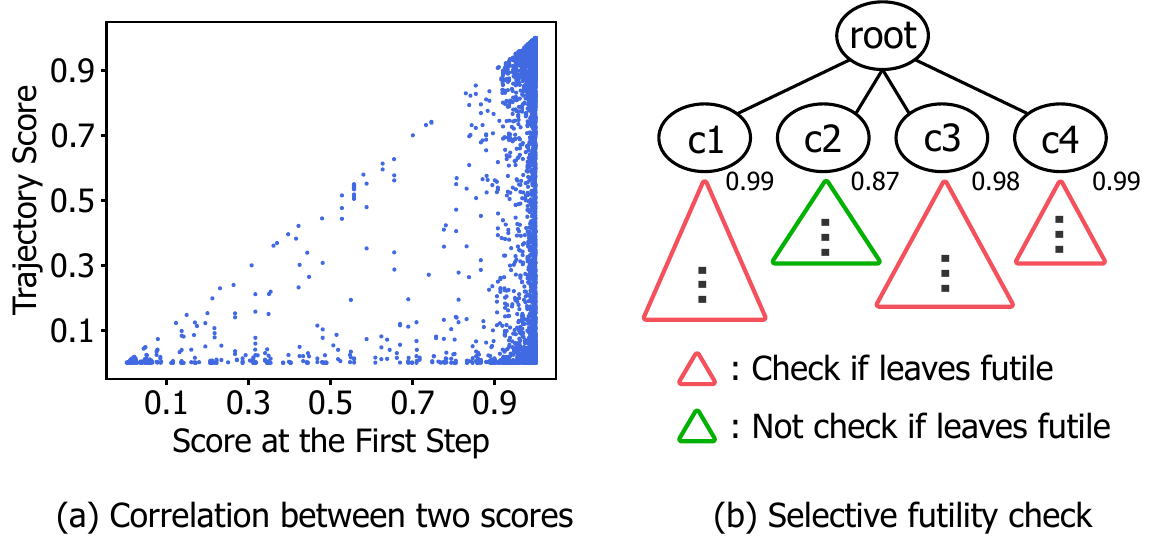}
\caption{Selective Futility Check Mechanism. (a) Trajectories with low initial scores tend to yield low final scores. (b) Selective futility check leveraging this correlation.}
\label{fig:negexit-check}
\end{figure}

Building on the observations in Section~\ref{subsec:motiv_negative_exit}, we can prematurely terminate the search process when all leaf nodes in the current search tree are identified as \textit{futile}.  
However, this strict condition for termination is seldom met in practice, leading to a limited invocation of early exits—approximately 5\% on the Math500 dataset. To further enhance resource efficiency, we introduce a relaxed criterion for early pruning based on the predictive value of early-stage rewards. Empirical observations suggest that trajectories with low initial rewards at the first depth tend to fall short of the acceptance threshold in their final scores.

As shown in Figure~\ref{fig:negexit-check} (a), the first-step reward exhibits a clear correlation with the final trajectory score, with trajectories that receive low rewards at the first depth rarely reaching the acceptance threshold later. 
Motivated by this trend, we exclude such low-potential branches from the negative-exit check, preventing them from delaying termination while still allowing the search procedure itself to proceed normally on potentially promising parts of the tree. 
An example of this selective futility check is shown in Figure~\ref{fig:negexit-check}(b).

\subsection{Parallelism Degree Control Policy}
We employ a dynamic Parallelism Degree Control Policy to optimize the balance between individual task latency and overall system throughput. 
This policy adaptively modulates the Degree of Parallelism (DOP) for each task by synthesizing two orthogonal factors: temporal seniority and score-based urgency. 
The primary objective is to prevent task starvation while simultaneously accelerating the clearance of tasks that are nearing completion, thereby optimizing the utilization of the running queue. Hereafter, \textbf{boosting} refers to executing multiple rollouts for a single task in parallel.

The scheduler calculates a parallelism degree score $S(i, t)$ for task $i$ at time $t$ to determine its resource allocation tier. 
The scoring function is defined as:
$$S(i, t) = \log(1 + \Delta t_i) + \beta \cdot I(\rho_i > \tau)$$
In this formulation, $\Delta t_i$ represents the elapsed time since the task’s arrival, calculated as $t_{current} - t_{arrival}$. 
The term $\log(1 + \Delta t_i)$ introduces a logarithmic scaling to the waiting time, which provides high initial sensitivity to ensure new tasks receive adequate resources while damping the score of long-running tasks to prevent resource monopolization.
The second term employs an indicator function $I(\rho_i > \tau)$, where $\rho_i$ is the progress ratio defined by the best intermediate score relative to the acceptance threshold, and $\tau$ is a proximity constant set to 0.9.

This 90\% Proximity Rule identifies tasks that are on the verge of satisfying the positive early exit criteria. 
When a task reaches this critical threshold ($\rho_i > 0.9$), the $\beta$ coefficient significantly boosts its priority score, triggering an immediate escalation in its parallelism degree. 
By concentrating computational power on these near-complete tasks, the system effectively minimizes the tail latency and rapidly frees up resources occupied by running tasks.
The resulting score $S(i, t)$ is then mapped to discrete parallelism degree, ensuring a balanced resource allocation for the active task set.

\subsection{Scheduling Algorithm}

\begin{algorithm}
\caption{Dynamic Scheduling with Boosting}
\begin{algorithmic}[1]
    \STATE \textbf{Phase 1: Job Admission}
    \STATE Admit pending jobs to $\mathcal{Q}_{run}$ such that $|\mathcal{Q}_{run}| \le M$, where $M$ is max concurrency.

    \vspace{2ex}
    \STATE \textbf{Phase 2: Scoring and Resource Allocation}
    \FORALL{$J_i \in \mathcal{Q}_{run}$}
        \STATE Compute Parallelism Degree Score $S_i$.
        \STATE Determine target parallelism $\hat{P}_i$:
        \STATE $\hat{P}_i = \begin{cases} 1 & \text{if } N_{obs}(J_i) < \tau \\ \max\left(1, \lfloor \frac{S_i}{\sum S_j} \cdot M \rfloor \right) & \text{otherwise} \end{cases}$
    \ENDFOR
    \STATE Finalize $P_i$ by allocating remaining budget $\mathcal{B}$ to jobs in descending order of $S_i$.

    \vspace{2ex}
    \STATE \textbf{Phase 3: Execution and Preemption}
    \FORALL{$J_i \in \mathcal{Q}_{run}$}
        \STATE Let $A_i$ be the number of active rollouts for $J_i$.
        \IF{$A_i > P_i$}
            \STATE \textbf{Preempt:} Cancel $(A_i - P_i)$ rollouts with minimum cumulative rewards.
        \ELSIF{$A_i < P_i$}
            \STATE \textbf{Expansion:} Launch $(P_i - A_i)$ new rollouts.
        \ENDIF
    \ENDFOR
\end{algorithmic}
\end{algorithm}

To maximize search efficiency under constrained computational budgets, we implement a dynamic scheduling policy that adaptively redistributes rollout resources among concurrent reasoning \textbf{jobs}, each corresponding to a single tree search. The algorithm begins by managing job admission, where pending jobs are moved into the active running queue only when the total rollout occupancy is below the maximum system capacity. 
This ensures a baseline level of progress for all admitted jobs while preventing resource exhaustion. 

Once jobs are active, the scheduler determines the optimal parallelism for each job based on its Parallelism Degree Score S. A key feature of this allocation strategy is the implementation of a gating mechanism for early-stage jobs; any job that has not yet surpassed a predefined observation threshold is restricted to serial execution. This prevents the premature commitment of heavy parallel resources to paths with insufficient data. For jobs beyond this threshold, the scheduler allocates the remaining rollout budget proportionally to their scores, prioritizing those with the highest urgency.

The final stage of the policy involves real-time execution adjustment to synchronize the actual number of running rollouts with these new assignments. The system achieves high elasticity by either launching additional rollouts to meet increased quotas or preempting existing ones when a job's allocation decreases. In the event of preemption, the scheduler intelligently terminates the least promising rollouts—those with the lowest cumulative rewards—to preserve the most valuable computations. This continuous feedback loop allows the system to remain highly responsive to the evolving search landscapes of multiple simultaneous jobs.

\section{Evaluation}
\label{sect:evaluation} 

\subsection{Methodology}

\paragraph{System Specifications and Models.}
All experiments were conducted on a single node equipped with four NVIDIA H100-SXM 80GB GPUs interconnected via NVLink. Two GPUs were allocated to generation model inference, while the remaining two GPUs were dedicated to reward model inference. We evaluate system performance under reasoning workloads using two generation models, Llama-3.1-8B-Instruct~\cite{llama3} and Qwen2.5-14B-Instruct~\cite{qwen2_5}, and conduct test-time inference scaling experiments independently for each model. For reward evaluation, we use Qwen2.5-Math-PRM-7B~\cite{qwen2_5_prm}. This configuration is used consistently across all experiments.

\paragraph{Datasets.}
We evaluate mathematical reasoning using Math500 and AMC23~\cite{math500,amc23}. Math500 is a 500-problem benchmark derived from MATH that measures step-by-step solution accuracy. AMC23 consists of problems from the 2023 American Mathematics Competitions and is used to evaluate generalization to unseen problems.
Due to the limited number of unique problems in AMC23, we augment the workload with distinct, accuracy-preserving prefixes only for performance measurements to fully utilize the system. All accuracy results are reported on the original AMC23 dataset, and prefix caching is enabled only within the MCTS process and disabled across requests by assigning unique prefixes per request.

\paragraph{Compared Systems.}
We compare the following system configurations. Positive early exit (PE) is adopted from prior work, while negative early exit (NE) and Boosting are our proposed extensions.
\begin{itemize}
    \item \textbf{Beam search}: A baseline with positive early exit applied to beam search, with the beam size fixed at 8 and the maximum number of concurrent jobs fixed at 16. We set the beam size to match the accuracy of Vanilla MCTS as closely as possible.
  \item \textbf{Vanilla}: A baseline system that serves generation models using MCTS-based test-time compute scaling, where tree search is performed sequentially for each request.
  \item \textbf{PE (Positive Early Exit)}: A system that adaptively terminates rollouts when intermediate reward signals indicate sufficient confidence, reducing unnecessary computation and allowing easy instances to complete quickly~\cite{fu2024certaindex}.
  \item \textbf{PE+NE (Negative Early Exit)}: An extension of PE that additionally stops rollouts early when all trajectories consistently exhibit low scores, identifying unlikely-to-be-solved instances early and reducing tail latency.
  \item \textbf{PE+NE+Boosting}: The full system configuration that utilizes the GPU computation freed by early exits to perform additional MCTS rollouts in parallel, improving both latency and throughput under higher load.
\end{itemize}

\subsection{End-to-end Latency Reduction}

\begin{figure}[t]
    \centering
\includegraphics[width=1.0\linewidth]{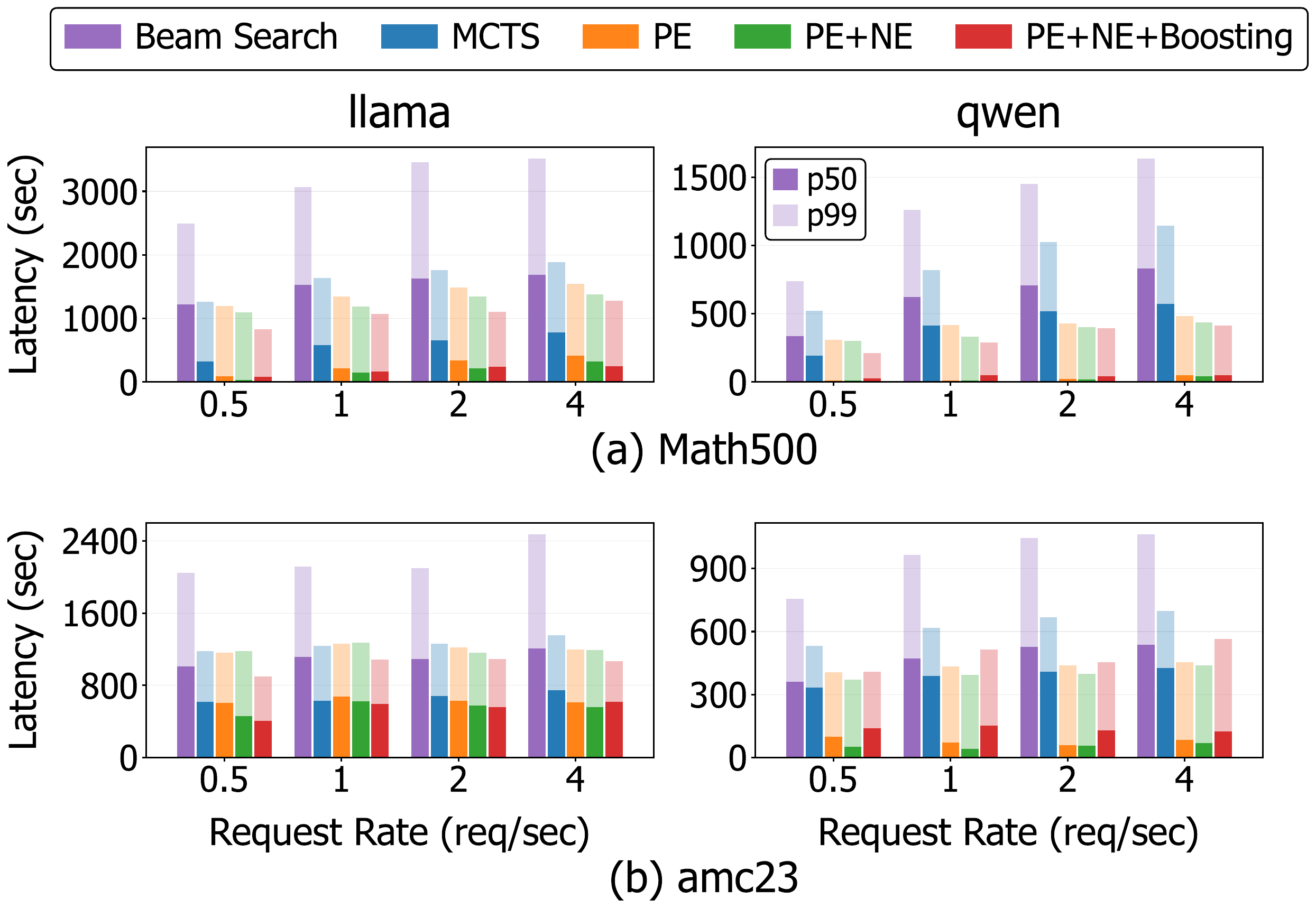}
    \caption{Impact of request arrival rates on p50 (darker) and p99 (lighter) end-to-end latency across different search strategies.}
    \label{fig:eval_main_1}
\end{figure}

Figure~\ref{fig:eval_main_1} shows the p50 and p99 end-to-end latency as the request arrival rate increases, illustrating how tail latency emerges in MCTS-based reasoning serving and how our techniques mitigate it. 
Beam search consistently exhibits the highest p99 latency, as its decoding cost scales quadratically with the beam size, allowing individual requests to monopolize GPU resources and induce severe queuing delays. 
Vanilla MCTS improves p99 latency by an average of 1.64$\times$ over beam search; however, because tree search is executed sequentially per request, long-running searches accumulate under higher load, resulting in persistent tail latency.

Positive Early Exit (PE) reduces unnecessary computation by terminating rollouts early when intermediate rewards indicate sufficient confidence, achieving an average p99 reduction of 1.38$\times$ relative to Vanilla. 
However, PE still fully explores low-promise requests, which can dominate the queue under high system pressure and limit its effectiveness in controlling tail latency. 
Negative Early Exit (NE) addresses this limitation by identifying search states whose cumulative scores cannot mathematically exceed a threshold and terminating them early. 
As a result, NE reduces p99 latency by 1.49$\times$ over Vanilla and provides an additional 1.08$\times$ improvement over PE, directly targeting tail behavior.

Boosting reinvests the GPU computation freed by early exits into parallel MCTS rollouts, accelerating request completion and alleviating queuing delays. 
The full configuration (PE+NE+Boosting) achieves an additional p99 reduction of 1.15$\times$ over PE and 1.07$\times$ over PE+NE. 
In the worst case, p99 latency decreases from 1,886~ms to 1,277~ms, corresponding to a 1.47$\times$ improvement over Vanilla and a 1.21$\times$ improvement over PE. 
One notable exception occurs in the (amc23, Qwen) configuration, where the p99 latency slightly increases compared to PE+NE. 
We attribute this to parallelization overhead; in this configuration most requests could be resolved with minimal sequential rollouts, but the scheduler’s aggressive parallelization unnecessarily monopolizes resources, thereby increasing queuing delays for concurrent requests.
Overall, these results show that tail latency in MCTS-based reasoning serving is a system-level resource management problem that requires combining early termination with parallelized test-time compute scaling.

\subsection{Request Throughput Improvement}
\label{subsec:accuracy}

\begin{figure}[t]
    \centering
\includegraphics[width=1.0\linewidth]{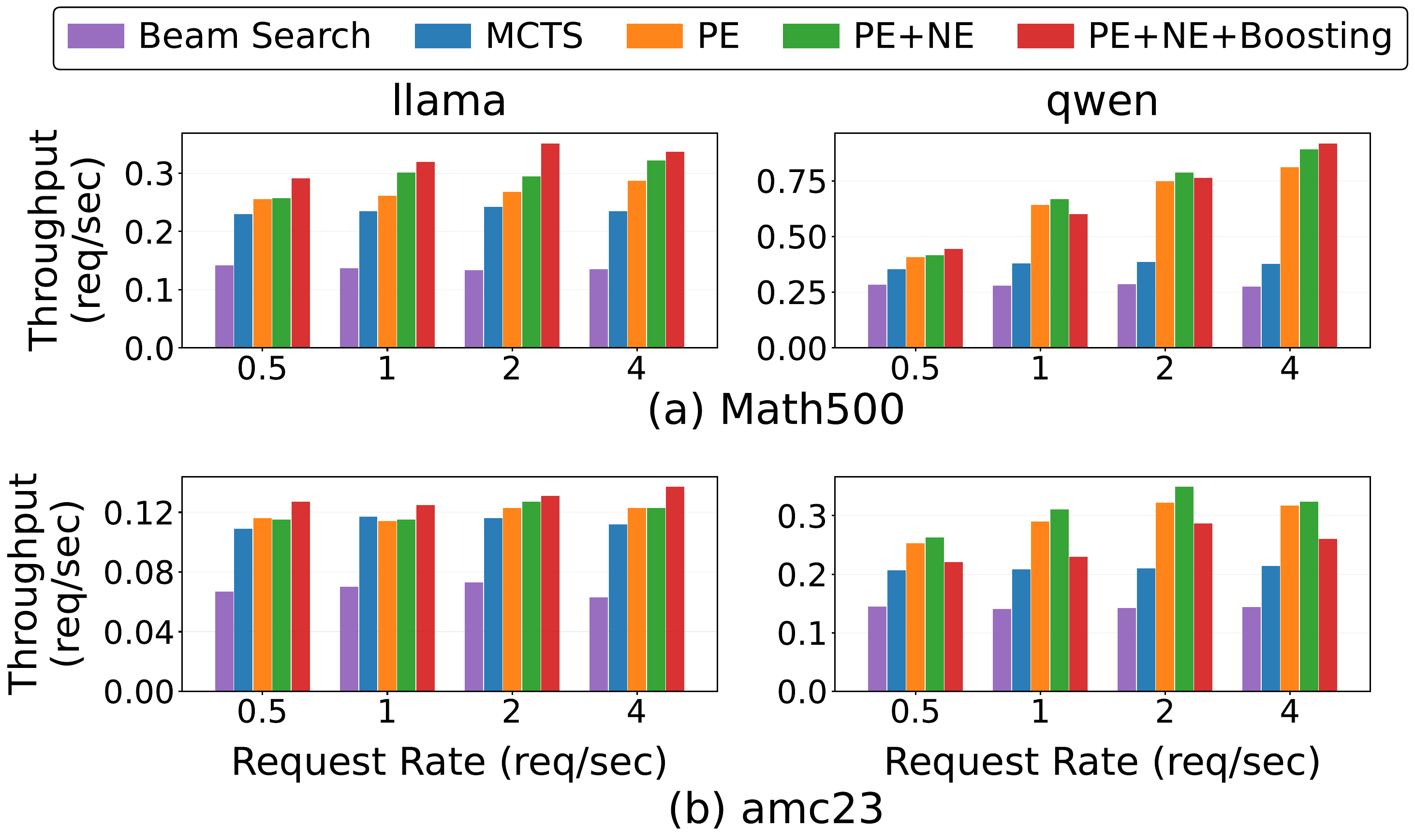}
    \caption{Request throughput across various model-task configurations. PE and NE consistently increase throughput. PE+NE+Boosting yields throughput gains across most cases.}
    \label{fig:eval_4}
\end{figure}

Figure~\ref{fig:eval_4} presents the request throughput across the evaluated systems. 
Similar to the latency trends, beam search exhibits the lowest throughput due to its heavy resource consumption. 
In contrast, Positive Early Exit (PE) and Negative Early Exit (NE) significantly reduce system overhead, yielding throughput improvements of $1.3\times$ and $1.36\times$ over the Vanilla MCTS baseline, respectively. 
The fully integrated system—incorporating Boosting—achieves throughput gains of $1.37\times$ for (Math500, Llama), $1.76\times$ for (Math500, Qwen), and $1.15\times$ for (amc23, Llama). 
These gains stem from the system’s ability to discover optimal solutions rapidly through refined search, allowing it to release computational resources back to the queue ahead of schedule and thus accommodate more concurrent requests.
However, in the (amc23, Qwen) configuration, applying Boosting on top of Negative Early Exit results in a throughput reduction to $0.8\times$ relative to NE alone. 
This behavior arises because most requests in this workload are solvable with a small number of serial rollouts, leaving limited opportunity for effective parallelization. 
As a result, the overhead of scheduling and coordinating parallel rollouts outweighs the marginal benefit of additional compute, leading to reduced throughput in this regime.

\subsection{Impact on Reasoning Accuracy}
\label{subsec:accuracy}


\begin{table}[!t]
\centering
\caption{Accuracy Comparison Across Optimizations (\%) }
\label{tab:accuracy}
\small 
\setlength{\tabcolsep}{5pt} 
\begin{tabular}{lcccc}
\toprule
\multicolumn{1}{c}{\textbf{Method}} & \multicolumn{2}{c}{\textbf{Math500}} & \multicolumn{2}{c}{\textbf{amc23}} \\
\cmidrule(r){2-3} \cmidrule(l){4-5}
& Llama & Qwen & Llama & Qwen \\
\midrule
Beam Search             & 72.9 & 87.1 & 52.5 & 67.5 \\
MCTS (Vanilla)          & 75.3 & 88.3 & 57.5 & 72.5 \\
MCTS + PE               & 76.2 & 88.3 & 47.5 & 69.2 \\
Ours (Full Suite)       & 74.8 & 87.6 & 55.0 & 65.0 \\
\bottomrule
\end{tabular}
\end{table}

Table~\ref{tab:accuracy} illustrates the impact of optimization on accuracy. 
In the Math500 dataset, the inclusion of PE yields a marginal accuracy improvement. We attribute this fluctuation primarily to the inherent randomness stemming from floating-point precision limits. Specifically, in highly parallelized GPU environments, the non-deterministic nature of floating-point accumulation—where subtle variations in the order of operations can lead to divergent rounding—appears to be the dominant factor. This sensitivity to numerical noise is notably more pronounced in the amc23 dataset, where the accuracy exhibits greater volatility. This is largely a function of dataset scale; whereas Math500 provides a more stable estimate across 500 instances, the smaller sample size of amc23 (40 instances) means that the outcome of a single request exerts a disproportionately large influence on the aggregate metric, thereby amplifying minor stochastic variations.

\section{Conclusion}

In conclusion, we show that the primary challenge in deploying MCTS-based test-time compute scaling is not reasoning accuracy, but managing its uneven computational cost under realistic serving workloads. By adopting a system-level perspective, we introduce Negative Early Exit to eliminate unproductive computation and Boosting to reallocate freed resources to higher-potential searches. Integrated into our system, these mechanisms preserve MCTS accuracy while reducing p99 end-to-end latency by up to 2.83$\times$ and improving throughput by up to 2.44$\times$, making test-time compute scaling practical for large-scale reasoning-serving systems.

\bibliography{references}
\bibliographystyle{icml2026}

\newpage
\appendix
\onecolumn
\section{Detailed Procedure of MCTS}
\label{sec:appendix_mcts_detail}

\begin{figure*}[t]
    \centering
    \includegraphics[width=1.0\linewidth]{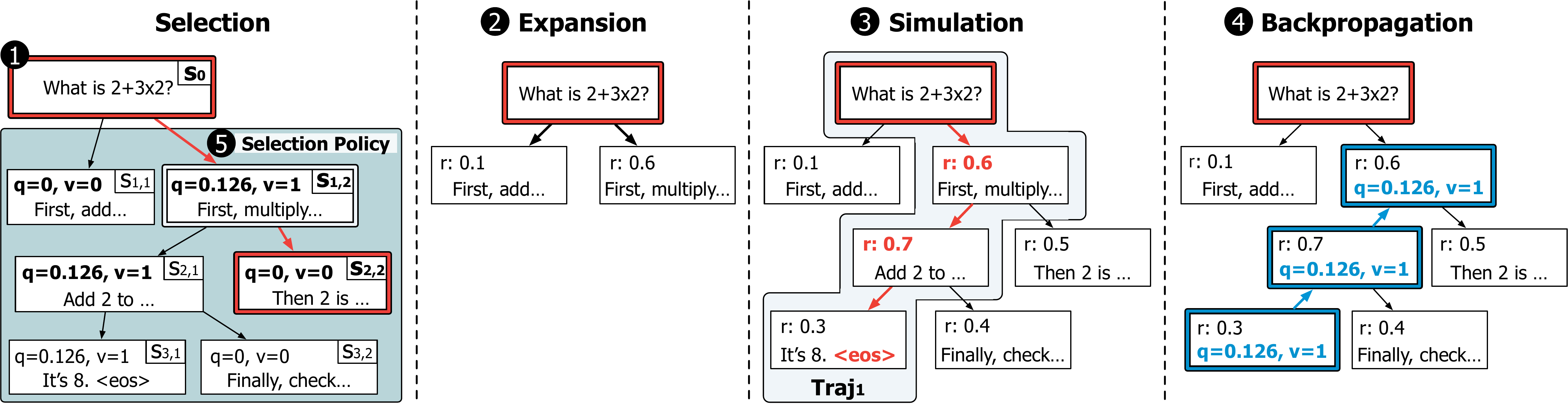}
    \caption{Node-reusing MCTS algorithm with four phases.}
    \label{fig:background_mcts_algorithm}
\end{figure*}

MCTS invests more computation time compared to other TTCS methods such as beam search or best-of-n, achieving higher accuracy ~\cite{ding2025dynamic}. Due to its consistently superior accuracy, recent work within TTCS has actively explored MCTS-based approaches~\cite{hao2023rap, feng2023tsllm, guan2025rstar_math, misaki2025ab_mcts, zhou2024lats, antoniades2024swe_search, shi2025mc_dml, ma2024repounderstander}. This is because, in domains with objective feedback, MCTS is particularly effective at leveraging external feedback as reward signals to select and refine promising solutions.
In MCTS-based reasoning inference, the algorithm incrementally grows a search tree for a given \reasoningreq, treating each partial token sequence as a node that captures an intermediate reasoning step. We refer to the input prompt as a \textbf{\reasoningreq}; to generate the output for a given \reasoningreq, an MCTS-based engine issues at every reasoning step a sequence-token generation query to the language model, and we call each such query a \textbf{\genreq}. From each node, the model applies alternative actions—LLM step generation in reasoning inference—to spawn multiple child nodes; repeating this process gradually expands the tree. A trajectory from the root to a terminal node constitutes a complete reasoning result.

To expand the search tree and select the final reasoning trajectory, a quantitative evaluation metric is required to measure each node's contribution to optimal reasoning. Accordingly, a reward model is employed, which assigns a score to each node based on its plausibility and consistency. While many variants of MCTS exist for reasoning inference, we target the \textit{node-reusing MCTS} employed across recent frameworks.~\cite{hao2023rap, feng2023tsllm,guan2025rstar_math, zhou2024lats, ma2024repounderstander}
Both node-reusing MCTS and traditional MCTS iterate four phases: (1) selection, (2) expansion, (3) simulation, and (4) backpropagation as shown in Figure~\ref{fig:background_mcts_algorithm}. In node-reusing MCTS, these phases proceed as follows.
\begin{description}[leftmargin=0pt,topsep=7pt,itemsep=3pt,partopsep=0pt]
    \item[Selection]
    Beginning from the root node $s_0$, the algorithm follows the selection policy until reaching the leaf node (\circled{1}, \circled{5}). There are two selection policies: \emph{Upper Confidence bounds applied to Trees} (UCT)~\cite{kocsis2006uct} and \emph{Predictor-UCT} (PUCT)~\cite{rosin2011puct}. In both policies, the child at step $t$ is chosen according to the following: 
    \begin{equation} \label{eq:uct} \text{Node}_{\text{selected}}=\arg\max_{s_{t,i}}\bigl(Q(s_{t,i})+U(s_{t,i})\bigr) \end{equation}
    Here, $Q(s_{t,i})$ denotes the exploitation term—how promising $i$-th child node $s_{t,i}$ appears at step $t$—and $U(s_{t,i})$ denotes the exploration term that favors under-explored children. Exploitation uses the average return over trajectories that traverse this child; accordingly, it is computed as the node’s \textit{value estimate}  ($q$ in \circled{5}) divided by its \textit{visit count} ($v$ in \circled{5}), whereas exploration uses \textit{visit count}. PUCT further scales the exploration term by the network’s prior policy, steering under-explored choices toward nodes the network deems promising and thus improving search efficiency. Accordingly, whereas traditional MCTS employs UCT, we adopt PUCT to concentrate the exploration on more promising nodes.
    \item[Expansion] 
    The search tree is expanded by generating child nodes from the leaf node selected during the selection (\circled{2}).
    \item[Simulation] 
    From the newly created node in the expansion, the algorithm iteratively generates the child nodes and selects one with the highest reward at each step, continuing until it reaches a terminal node (\circled{3}).
    \item[Backpropagation]
    When the simulation produces a terminal node that ends with an \texttt{eos} token, a trajectory is completed. We then backpropagate along this trajectory, adding the trajectory’s cumulative reward to each node’s \textit{value estimate} while incrementing its \textit{visit count} by one (\circled{4}). Whereas traditional MCTS discards nodes created during simulation, node-reusing MCTS retains them and proceeds to the next selection step as shown in \circled{5}.
\end{description}
In node-reusing MCTS, each repetition of the four phases yields one complete trajectory, called a \textit{rollout}. After generating a fixed number of trajectories, the trajectory with the highest cumulative reward is output. Since a single rollout yields one trajectory and the next rollout starts afterward, trajectories are generated sequentially. We target node-reusing MCTS; unless otherwise noted, all subsequent references to MCTS denote node-reusing MCTS.

\section{Detailed Procedure of Beam Search}
\label{sec:appendix_bs_detail}


\begin{figure}
    \centering
    \includegraphics[width=1.0\linewidth]{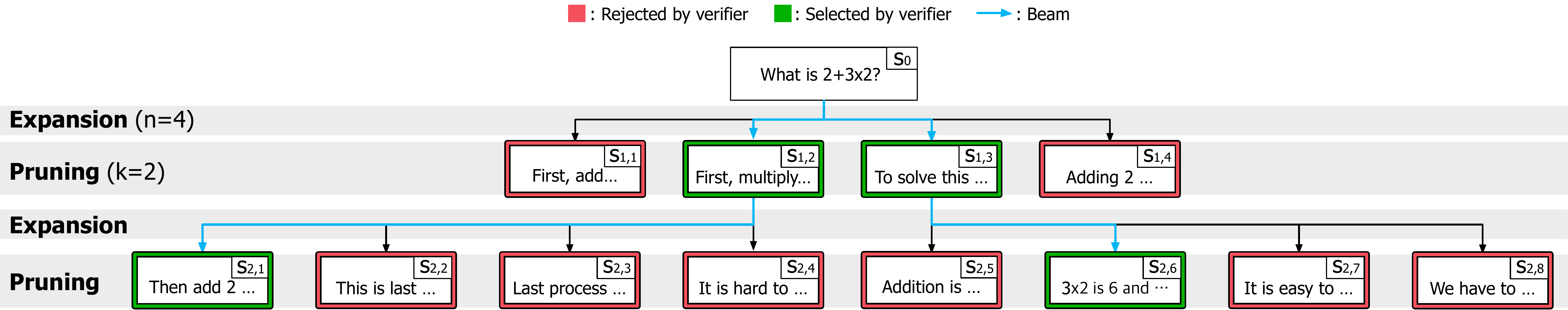}
    \caption{Beam Search algorithm with beam width 2 and 4 sampled candidates per beam.}
    \label{fig:appendix_bs_algorithm}
\end{figure}

Beam Search is a parallel search method that maintains a small set of high-scoring partial solutions while incrementally extending them over a fixed number of reasoning steps. In the context of multi-step reasoning, the term \textbf{beam} typically refers to a partial trajectory, i.e., a prefix of a reasoning chain up to the current time step. The algorithm proceeds in discrete steps. At each step, it expands the current set of beams, scores the resulting candidates with a verifier, and prunes to retain only the top $k$ candidates, where $k$ is a predefined hyperparameter called the \textbf{beam width}.
In summary, as illustrated in Figure~\ref{fig:appendix_bs_algorithm}, Beam Search proceeds by repeatedly alternating between two phases: (1) expansion and (2) pruning.
\begin{description}[leftmargin=0pt,topsep=7pt,itemsep=3pt,partopsep=0pt]
    \item[Expansion]
    At time step $t$, each of the $k$ surviving beams is expanded by generating candidate next reasoning steps conditioned on its current partial trajectory. For each beam, $n$ candidate steps are generated, where $n$ is a predefined hyperparameter. As in MCTS, a reasoning step corresponds to a multi-token sequence.
    \item[Pruning]
    After expansion, the $nk$ candidates are reduced to $k$ surviving beams, and the rest are discarded. This step maintains a fixed beam width of $k$ at every time step. Concretely, the top-$k$ candidates are selected deterministically based on their accumulated scores, which are computed using an external verifier.
\end{description}

The algorithm repeats the expansion--pruning phases until a termination criterion is met, typically a maximum reasoning depth $T$ (i.e., a fixed number of reasoning steps). After termination, Beam Search selects the final solution among the completed trajectories. The output is chosen as the completed beam with the highest accumulated verifier score.


\end{document}